\documentclass{svproc}

\usepackage[T1]{fontenc}
\usepackage[utf8]{inputenc}

\usepackage{color}
\usepackage[caption=false]{subfig}
\usepackage{times}
\usepackage{epsfig}
\usepackage{wrapfig}
\usepackage{amsmath}
\usepackage{amssymb}
\usepackage{algorithmicx}
\usepackage{algpseudocode}
\usepackage[ruled,vlined,linesnumbered]{algorithm2e}
\usepackage[misc]{ifsym}
\usepackage{tabularx}
\usepackage{multirow}
\usepackage{hyphenat} % for preventing hyphenation, using environment \nohyphens{(text block)}
\usepackage{graphbox}

\usepackage{soul}
\usepackage{url}
\usepackage{fancybox}
\usepackage[export]{adjustbox}
\usepackage{diagbox}
\usepackage[colorlinks]{hyperref}
\graphicspath{{figures/}}

\usepackage{bookmark}
\makeatletter
\def\toclevel@title{-1}
\def\toclevel@author{0}
\makeatother

\begin{document}
\mainmatter              % start of a contribution
\title{DGORL: Distributed Graph Optimization based Relative Localization of Multi-Robot Systems}

\titlerunning{Distributed Graph Optimization based Relative Localization}  % abbreviated title (for running head)
%                                     also used for the TOC unless
%                                     \toctitle is used
%
\author{Ehsan Latif and Ramviyas Parasuraman}
\authorrunning{Latif and Parasuraman} % abbreviated author list (for running head)
%
%%%% list of authors for the TOC (use if author list has to be modified)
%\tocauthor{Ivar Ekeland, Roger Temam, Jeffrey Dean, David Grove, Craig Chambers, Kim B. Bruce, and Elisa Bertino}
%

\institute{School of Computing, University of Georgia, Athens, GA 30602, USA. \\
email: \email{ehsan.latif@uga.edu, ramviyas@uga.edu} \\
Codes \url{https://github.com/herolab-uga/DGORL}
}

\maketitle              % typeset the title of the contribution

\begin{abstract}
An optimization problem is at the heart of many robotics estimating, planning, and optimum control problems. Several attempts have been made at model-based multi-robot localization, and few have formulated the multi-robot collaborative localization problem as a factor graph problem to solve through graph optimization. Here, the optimization objective is to minimize the errors of estimating the relative location estimates in a distributed manner. Our novel graph-theoretic approach to solving this problem consists of three major components; (connectivity) graph formation, expansion through transition model, and optimization of relative poses. First, we estimate the relative pose-connectivity graph using the received signal strength between the connected robots, indicating relative ranges between them. Then, we apply a motion model to formulate graph expansion and optimize them using g$^2$o graph optimization as a distributed solver over dynamic networks. Finally, we theoretically analyze the algorithm and numerically validate its optimality and performance through extensive simulations. The results demonstrate the practicality of the proposed solution compared to a state-of-the-art algorithm for collaborative localization in multi-robot systems.
%The objective to be maximized or minimized in most of these optimization problems is made up of several local components or terms, i.e., they depend on just a tiny fraction of the variables. 
%We utilize the graph theory to solve this well-defined problem of multi-robot relative localization. 
\end{abstract}

\section{Introduction}
\label{sec:intro}
The estimation of a relative pose, including position and orientation, \cite{islam2021robot}, for multi-robot systems (MRS) \cite{xianjia2021applications} is the foundation for higher-level tasks like collision avoidance, cooperative transportation, and object manipulation. Motion capture systems \cite{najafi2019adaptive}, ultra-wideband (UWB) systems with anchors, and RTK-GPS systems are a few examples of established multi-robot relative positioning solutions that currently rely on the deployment of physical anchor or base stations in the application. These plans, however, are not suitable for large areas or interior settings where it is difficult to convert the infrastructure, which limits the overall performance and application possibilities of multi-robot systems and makes their use more difficult and expensive.
Furthermore, extraction of range and bearing measurements from cameras and visual makers, while another practical approach, has the drawbacks of having a small field of vision, a short-range, obscured by nearby objects, and maybe requiring much computational power. The use of distance measurements from sensors like radars, Lidars, and UWB to achieve relative localization, on the other hand, has recently attracted more significant interest.

The multi-robot relative localization (MRL) problem, which refers to detecting and locating the relative configurations of mobile agents (typically with fewer sensor data such as relative range or bearing) concerning other agents or landmarks, is critical in MRS because it is required for robot teaming and swarming \cite{guo2017ultra,fink2012distributed}. 
As a result, many applications are frequently confronted with the relative localization problem, including formation control, cooperative transportation, perimeter surveillance, area coverage, and situational awareness. 
The relative localization and mapping (aka multi-robot SLAM) is an extension of the MRL problem. While several researchers have proposed novel solutions to the multi-robot map merging problem using pose graph matching and optimization techniques, they rely on extensive sensor data inputs (such as point clouds or Lidar scans) \cite{dube2017online,mangelson2018pairwise,tian2022kimera}. Therefore, solving the MRL problem with relative range or bearing in a distributed manner is desirable and scalable in MRS \cite{latif2022multi}.

\begin{figure}[t]
    %\centering
%\begin{subfigure}{\linewidth}
\centering
\includegraphics[width=0.95\textwidth]{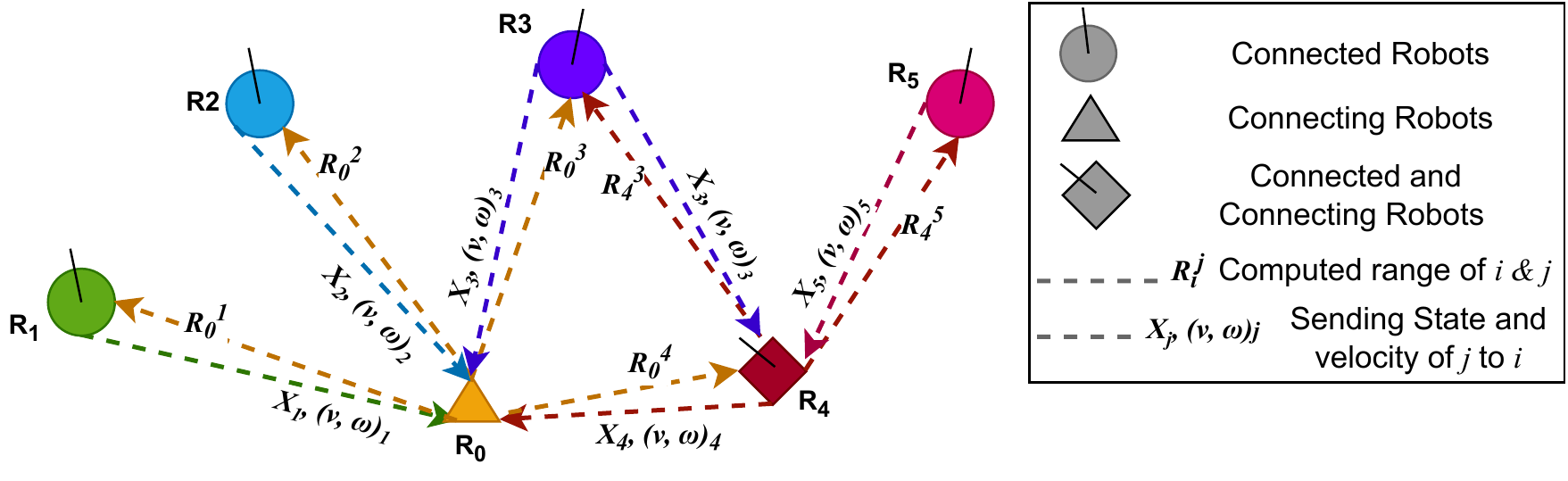}
%\end{subfigure}
\caption{Overview of configuration space of a multi-robot system, sharing their pose ($x_i$) and relative range ($R^j_i$) measurements in our DGORL solution.}
    \label{fig:overview}
\end{figure}

\textcolor{black}{Distributed optimization is the problem of minimizing a joint objective function that is the sum of many local objective functions, each corresponding to a computer node. 
We can model many fundamental activities in this area as distributed optimization problems, which have significant implications for multi-robot systems. Examples include cooperative estimation \cite{9197241}, multiagent learning \cite{wai2018multi}, and collaborative motion planning. The distributed optimization formulation provides a versatile and effective paradigm for creating distributed algorithms for numerous multi-robot problems.}

%Each of these problems can be represented by a graph. Each node in the graph represents a state variable that needs to be optimized, whereas each edge linking two variables is a paired observation of the two nodes it connects. In the literature, various tactics have been proposed to address this set of problems. A straightforward implementation using well-known methods like Gauss-Newton, Levenberg-Marquardt (LM), Gauss-Seidel relaxation, or different iterations of gradient descent typically produces satisfactory results for most applications. However, extensive effort and subject-matter knowledge are required to achieve the highest outcome.

In consumer electronics, Wi-Fi is one of the most extensively utilized wireless technology for indoor wireless networks. 
The ubiquitous availability of Received Signal Strength Indicator (RSSI) measurement on such inexpensive commercial devices is the RSSI measured from an Access Point (AP) or a Wireless Sensor Robot (WSN). The RSSI value can be used in various applications, including relative localization \cite{wang2019active,parashar2020particle,latif2022online}, cooperative control \cite{luo2019multi,parasuraman2019consensus}, and communication optimization \cite{parasuraman2018kalman,parasuraman2017new}.

In this paper, we formulate the MRL problem as a graph optimization problem and solve it in a distributed manner using a polynomial-time optimizer called the General Graph Optimization (g$^2$o \cite{kummerle2011g}). g$^2$o is an open-source graph-based framework to handle the nonlinear error problems and is used to optimize global measurement pose using the initial global measurement poses and local relative pose constraints. 

Our solution, termed DGORL, aims to achieve high localization accuracy efficiently in a distributed fashion. DGORL forms relative position-weighted connectivity graphs using RSSI as local sensor data then expands graphs based on possible positions at an instant and further optimizes to fetch relative position estimates for all connected robots.
See Fig.~\ref{fig:overview} for an overview of the configuration space of DGORL.

The main contributions of this paper are listed below.
\begin{enumerate}
    \item A novel distributed, efficient, and precise relative localization system based on shared inertial measurements and RSSI inputs from connected robots.
    \item Position-weighted connectivity graph construction and optimization strategy tailored specifically for obtaining reliable relative pose estimates.
    \item Theoretical and numerical analysis to evaluate the performance of the algorithm.
    \item Validation of accuracy and efficiency of the DGORL compared to the recent collaborative multi-robot localization algorithm \cite{wiktor2020ICRA}, which used covariance intersection technique to address the temporal correlation between received signals.
    \item Open-sourcing of the codes\footnote{\url{http://github.com/herolab-uga/DGORL}} for use and improvement by the research community.
\end{enumerate}

\section{Related Work}
\label{sec:literature}
Most recent solutions to the simultaneous localization and mapping (SLAM) and MRL problem are based on graph optimization (i.e., all robot poses and landmark positions compose the graph's nodes, while each edge encodes a measurement constraint) \cite{kummerle2011g}. A conventional graph formulation, on the other hand, may suffer from unbounded processing and memory complexity, which might constantly expand over time. This is because new robot poses (and new landmarks in the case of feature-based SLAM) are constantly being added to the graph, resulting in an increase in the number of nodes over time; additionally, if frequent loop-closing events occur in SLAM, loop-closure constraints (edges) can significantly increase the graph density \cite{6698835}. For example, this could be the case if a service robot works for an extended time inside an office building.

Particularly, graph optimization and factoring have been recently proposed in the literature to solve different variants of the MRL problem \cite{zheng2022multi,HAO2022105152,sahawneh2017factor}.
Even though the issue of reducing the complexity of graph optimization has recently been addressed \cite{carlevaris2013generic,johannsson2013temporally}, to the best of our knowledge, little work has yet explicitly taken into account estimation consistency (i.e., unbiased estimates and an estimated covariance more significant than or equal to the actual covariance \cite{bar2004estimation}) in the design of graph reduction (sparsification) schemes. This is a critical flaw because if an estimator is inconsistent, the accuracy of the derived state estimations is unclear, making it untrustworthy \cite{indelman2014multi}.
Moreover, the performance and efficiency of approaches to the localization problem in dynamic environments are significantly traded off.

Most cooperative localization methods entail robot communication and observation, which makes any step prone to inaccuracy. 
In a recent attempt at multi-robot localization, many robots can locate themselves jointly using Terrain Relative Navigation (TRN) \cite{wiktor2020ICRA}. The localization estimation utilizing shared information fusion has been improved by using an estimator structure that takes advantage of covariance intersection (CI) to reduce one source of measurement correlation while properly including others. 
Similarly, a work \cite{chang2021resilient} developed a CI-based localization method with an explicit communication update and guaranteed estimation consistency simultaneously would increase the robustness of multi-robot cooperative localization methods in a dispersed setting. However, robots keep a system-wide estimate in their algorithm, which relative observations can instantly update. Consequently, it increases the computational complexity at a centralized server and over-burdened the robot to keep track of dynamics for global positioning accurately.
Unlike the explicit CI modeling methods, our objective is to localize the robot relative to other robots in a distributed fashion by utilizing the polynomial-time graph optimization technique with high accuracy and efficiency.

Therefore, this paper proposes a graph-based optimization algorithm to address the gaps mentioned above. To form a graph using shared RSSI information among robots, we employ a Relative Pose Measurement Graph (RPMG) using observability analysis \cite{HAO2022105152}. Once we have created a connected, reliable graph, we exploit particle filtering over the motion model to expand the graph based on mobility constraints. Finally, we construct a k-possible combination of graphs for each robot which needs to be optimized. We use the polynomial-time graph optimizer (g$^2$o \cite{kummerle2011g}) for graph optimization with a distributed constraint optimization.

\section{Problem Formulation and the Proposed DGORL Solution}
\label{sec:problem}
An overview of the typical MRS components is given in this section, along with a thorough explanation of the suggested distributed graph optimization formulation.

\textbf{Multi-Robot System}
Robotic members of an MRS are divided into disjoint (isolated/disconnected from others) and connected (operating collaboratively) robots, which can be either ground-based or aerial. The robot that enters the measuring range of the monitoring robot at any given moment is referred to as the observed robot. A robot that takes measurements of a random robot, among other robots, is the observation robot. The following qualities are presumptive for the considered MRS's robotic members:
\begin{itemize}
    \item Wireless communication is used to share information across the MRS.
     \item The watching robot can extract the neighboring robot's relative range (e.g., through RSSI measurements) and can uniquely identify each robot in its field of view.
    \item While the observed robots have limited sensory and computational capabilities, the observing robot may use high-precision sensors to carry out its self-localization.
    %\item Robots can communicate with one another using various tools and protocols.
    \item We restrict the movement of the robots within a two-dimensional planar space.
\end{itemize}

Assume at a given time $t$, a team of robots contains $n \in \mathbb{N}$ connected robots can form a weighted undirected graph, denoted by $G=(V,E,A)$, of order $n$ consists of a vertex set $V=\{v_1,...,v_n\}$, an undirected edge set $E \in V \times V$ is a range between connected robots and an adjacency matrix $A=\{a_{ij}\}_{n \times n}$ with non-negative element $a_{ij}>0$, if $(v_i,v_j) \in E$ and $a_{ij}=0$ otherwise. 
An undirected edge $e_{ij}$ in the graph $G$ is denoted by the unordered pair of robots $(v_i,v_j)$, which means that robots $v_i$ and $v_j$ can exchange information with each other. 

Here, we only consider the undirected graphs, which indicates that the communications among robots are all bidirectional. Then, the connection weight between robots $v_i$ and $v_j$ in graph $G$ satisfies $a_{ij}=a_{ji}>0$ if they are connected; otherwise, $a_{ij}=a_{ji}=0$. Without loss of generality, it is noted that $a_{ii}=0$ indicates no self-connection in the graph. The degree of robot $v_i$ is defined by $d(v_i) = \sum_{j=1}^{n} a_{ij}$ where $j \neq i$ and $i=1,2,...,n$. The Laplacian matrix of the graph $G$ is defined as $L_n=D-A$, where $D$ is the diagonal with $D=diag\{d(v_1),d(v_2),...,d(v_n)\}$. If the graph $G$ is undirected, $L_n$ is symmetric and positive semi-definite. A path between robots $v_i$ and $v_j$ in a graph $G$ is a sequence of edges $(v_i,v_{i1}),(v_{i1},v_{i2}),...,(v_{ik},v_j)$ in the graph with distinct robots $v_{il} \in V$. An undirected graph $G$ is connected if there exists a path between any pair of distinct robots $v_i$ and $v_j$ where $(i,j=1,...,n)$.

\begin{figure}[t]
%\begin{subfigure}{\linewidth}
\centering
%\hspace{-8mm}
\includegraphics[width=0.99\linewidth]{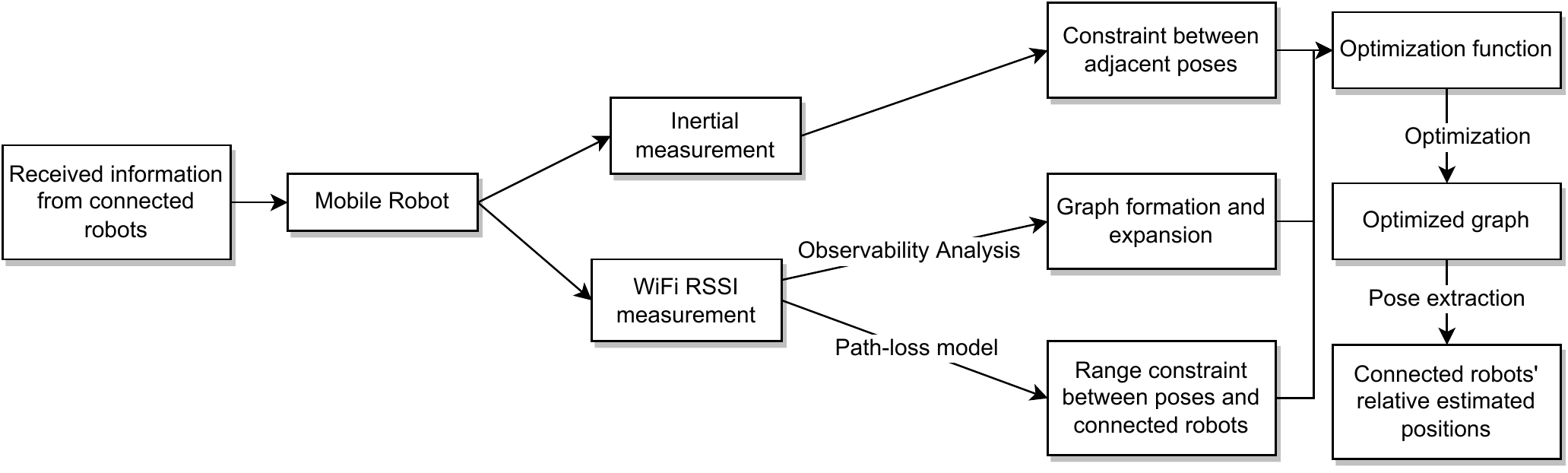}
%\end{subfigure}
\caption{The Distributed Graph Optimization for Relative Localization of Multi-Robot Systems (DGORL) system architecture shows input (Received connected robot's motion information and RSSI) and output (relative pose estimation) for robot $i$. The robot generates and expends graphs based on observability for optimization along with the constraints set up through local and received inertial measurement, which further fed into the optimization function. The optimized graph yields relative pose estimates for all connected robots.}
    \label{fig:architecture}
\end{figure}

In this paper, we formulated a multi-robot system as a graph problem to find the optimized solution for relative localization based on inequality constraints with a time-varying domain. Furthermore, we have segmented the solution into three different components, which can be solved simultaneously for every iteration over distributed constraints: 1) graph formation, 2) expansion through transition, and 3) optimization. See Fig.~\ref{fig:architecture} for sequential procedure of DGORL.

\subsection{Graph Formation}
An undirected graph $G = (V, E )$, where $(i, j) \in E$ if and only if $(i, j)$ and $(j, i)$ both are in $E$, is called the underlying
undirected graph of $G$. $G$ is connected if its underlying undirected
graph is connected. An isolated and connected subgraph of $G$ is
called a component.
Suppose that an undirected graph G has n nodes and m edges. The
incidence matrix of $G$, denoted by $A(G)$, is an $m \times n$ matrix whose
$m$ rows and $n$ columns correspond to the $m$ edges and $n$ nodes of
$G$, respectively. Each element $a_{ij}$ of $A(G)$ is defined as:

\[ a_{ij} =
 \begin{cases}
 1 \text{ if node } v_j \text{ is the tail/head of edge } e_i,\\
 0 \text{ otherwise}\\
 \end{cases}
 \]

The following lemma describes the relation between the rank of
$A(G)$ and the connectivity of $G$.\\
\textbf{Lemma 1}. Let $G$ be an undirected graph with $n$ nodes and $A(G)$
be its incidence matrix. If $G$ is composed of $\lambda$ components, the rank
of $A(G)$ satisfies
\[rank(A(G)) = n - \lambda \leq n - 1\]

The overall algorithm is mainly composed of two steps, as depicted in Algorithms \ref{alg: ERPMG} and \ref{alg: OCheck}, respectively.
The first step is to generate the Estimated Relative Position Measurement Graph (ERPMG); $G_E = (V_E, E_E, A_E )$ based on the Relative Position Estimation Graph (RPMG); $G = (V, E, A)$, which can be built-up using received range information from connected robots of an n-robot system, which describes the relative position measurements among robots. In this step, the node-set $V_E$ of the $ERPMG= G_E$ is initialized to be the same as that of the $RPMG= G$, and the edge set $E_E$ of the $ERPMG= {G_E}$ is initialized to be empty. Then, add all relative position measurement edges of $E$ into the edge set $E_E$ if edge $e_{ij}$ satisfy the motion constraint; $J^{F_j}V_{B_j} \neq 0$. A concise description of this step is illustrated in Algorithm \ref{alg: ERPMG}.

 \begin{algorithm}[ht]
 \SetAlgoLined
  Input: State x, input u, RPMG G = (V, E, A)\;
  Output: ERPMG $G_E$\;
  Initialize ERPMG $G_E = (V_E , E_E )$ with $V_E = V$, $E_E = \phi$ \;
  \For{ each node $v_i \in V$ d}{
      \For{ each edge $e_{ij} \in E$}{
            % add edge $e_{ij}$ to $E_E$ \;%add condition before adding new edge
            \If{$J^{F_j}V_{B_j} \neq 0$}{
                add edge  ${e}_{ij}$ to $E_E$ \;%update for single edge
            }
      }
  }
  \caption{ERPMG Formation}
  \label{alg: ERPMG}
\end{algorithm}

The second step (See Alg.~\ref{alg: OCheck}) is to examine the observability of the n-robot system according to the $ERPMG= {G_E}$ generated by Alg.~\ref{alg: ERPMG}. We initialize the incidence matrix $A(G_E)$ and the diagonal matrix $L_n$ to be zero matrices, with their dimensions determined by the number of nodes and edges in $G_E$. Then construct the incidence
matrix $A(G_E)$ and the diagonal matrix $L_n$. The incidence matrix $A(G_E)$ describes which nodes are connected by which edges and is constructed based on the topology of $G_E$. The diagonal matrix $L_n$ describes the weight of the edges in $G_E$ with the weight vector of each edge. The edges in $A(G_E)$ and $L_n$ take the same order. Then, we can obtain the spectral matrix $C(G_E)$ based on $A(G_E)$ and $L_n$. The observability of the n-robot system can be determined by evaluating the rank of $C(G_E)$.

 \begin{algorithm}[ht]
 \SetAlgoLined
  Input:  n = $|V_E|$, m = $|E_E|$ \;
  Initialize the incidence matrix: $A(G_E) = m \times n$\;
  Initialize the diagonal matrix: $L_n = n \times n$ \;
  k = 0\;
  \For{ each node $v_i \in V_E$}{
      \For{ each edge $(i, j) \in E_E$}{
            $k = k + 1$\;
            $A(G_E )[k, i] = 1$\;
            % $A(G_E )[k, j] = -1$\;
            $W[3k - 2 : 3k, 3k - 2 : 3k] = diag(L_n((i, j)))$\;
      }
   }
   \If{ rank ($C(G_E)$) is equal to 4$(n - 1)$ }{
        return True\; % Observable;
    }
   \Else{
        return False\; % Unobservable;
    }
  \caption{Observability Checking}
  \label{alg: OCheck}
\end{algorithm}

\subsection{Expansion through Transition}
Initial position and velocity constraints are known to each robot; hence locally constructed ERPMG tends to change based on relative motion in the network. We have limited the number of possible positions of an individual robot to $k$ by exploiting the concept of particle filtering. The motion process is carried out in the 2-D state-space $X_{n,t}$ includes position $(x_{n,t}, y_{n,t})$ and orientation $\phi_n, t$.\\
The robot model $f(*)$ can be written as:
\begin{equation}
\begin{aligned}
x_{n,t+1} &= x_{n,t} + v_{n,t}\Delta t cos(\phi_n,t)\\
y_{n,t+1} &= y_{n,t} + v_{n,t}\Delta t sin(\phi_n,t)\\
\phi_{n,t+1} &= \phi_{n,t} + \omega_{n,t}\Delta t
\end{aligned}
    \label{eqn: model}
\end{equation} 

In the Eq.~\eqref{eqn: model}, $v_{n,t}$ and $\omega_{n,t}$ are the velocity and angular velocity of the robot $n$ at time, $t$ respectively. $\delta t$ represents the time interval between two control inputs. Based on the
robot model $f(*)$ , the control input of the robot $n$ at time $t$ is defined as:
\begin{equation}
    u_{n,t} = [v_{n,t}, \omega_{n,t}]
    \label{eqn: ideal}
\end{equation} 
It is worth mentioning that Eq.~\eqref{eqn: ideal} is the ideal kinematic model of the robot.
Under pure rolling conditions, the robot's motion follows this equation. However, imperfections are unavoidable in real devices. Complex wheel-ground interactions and system noises cause disturbances to the robots. Moreover, these disturbances are modeled as Gaussian random variables, characterized by their mean and covariance matrices. Thus, the model of a given robot $n$ at time $t$ is defined as:
\begin{equation}
    x_{n,t+1} = f(x_{n,t} , u_{n,t}) + \mathcal{N}_{\text{noise}_{n,t}}
\end{equation} 

Here, $x_{n,t} \in R_{nx}$ and $u_{n,t} \in R_{nu}$ are the state and control input of the robot $n$ at the time, $t$ respectively. $\mathcal{N}_{\text{noise}} \in R_{nx}$ is an unknown disturbance with Gaussian probability distribution. Considering the disturbance level in the real environment, $\mathcal{N}_{n,t}$ noise in simulations is set to diag(0.1 m, 0.1 m, 0.5 deg).

As each robot also receives RSSI from other robots, we can find the intersecting region as an area of interest to map the estimated relative position for each robot using the model and find $k$ soft max out of them. Once we have $k$ possible positions of each robot, we can generate $<n^k$ solvable graphs for optimization.

\subsection{Optimization}
\subsubsection{Constraints}
We are interested in solving the constrained convex optimization problem over a multi-robot network in a distributed fashion. More specifically, we consider a network with $n$ robots, labeled by $V=\{1,2,...,n\}$ and $k$ possible connections to other robots. Every robot $i$ has a local convex objective function and a global constraint set. The network cost function is given by:
\begin{eqnarray} 
\mathrm {minimize} ~~&f( \mathbf {x}) = \sum \limits _{i=1}^{N} f_{i}( \mathbf {x}) ~~ \mathrm {subject~to} ~~& \mathbf {x}\in \mathcal {D}= \left \{{ \mathbf {x}\in \mathbf {R}^{k}~:~c( \mathbf {x}) \leq 0 }\right \}
\label{eqn: constraints}
\end{eqnarray}
Here, $x\in R^k$ is a global decision vector; $f_i: R^k\to R$ is the convex objective function of robot $i$ known only by robot $i$; $D$ is a bounded convex domain, which is (without loss of generality) characterized by an inequality constraint, i.e., $c(x)\leq 0$, where $c: R^k\to R$ is a convex constraint function. All the robots know it.
We assume that $D$ is contained in a Euclidean ball of radius $R$, that is:
\begin{equation} \mathcal {D}\subseteq \mathcal {B} = \left \{{ \mathbf {x}\in \mathbf {R}^{k}~:~\| \mathbf {x}\|_{2} \leq R }\right \}. \end{equation}
We also assume that there exists a point $\hat{x}$ such that the inequality constraint is strictly feasible, i.e., $c(\hat{x})<0$.
We introduce a regularized Lagrangian function to deal with the inequality constraint $c(x)$.
\begin{align} \mathsf {L}( \mathbf {x},\lambda )=&\sum \limits _{i=1}^{N} f_{i}( \mathbf {x}) + \lambda N c( \mathbf {x}) - \frac {\gamma }{2} N \lambda ^{2} \notag &=&\sum \limits _{i=1}^{N} \left [{ f_{i}( \mathbf {x}) + \lambda c( \mathbf {x}) - \frac {\gamma }{2} \lambda ^{2} }\right ] \notag \\=&\sum \limits _{i=1}^{N} \mathsf {L}_{i}( \mathbf {x},\lambda ) , \end{align}
where, we have replaced the inequality constraint $c(x)\leq 0$ with $N_c(x)\leq 0$ , and $\gamma>0$ is some parameter. It is noted that $L_i(x,\lambda)$ is known only by robot $i$.

\subsubsection{Model}
We consider a time-varying network model that has been widely considered in the literature \cite{6698835,kummerle2011g,solver}. The robots’ connectivity at time $t$ can be represented by an undirected graph $G(t)=(V,E(t),A(t))$ , where $E(t)$ is the set of activated edges at time $t$ , i.e., edge $(i,j)\in E(t)$ if and only if robot $i$ can receive data from robot $j$ , and we assign a weight $[A(t)]ij>0$ to the data on edge $(i,j)$ at time $k$ . Note that the set $E(t)$ includes self-edges $(i,i)$ for all $i$. We make the following standard assumption on the graph $G(t)$.

\textbf{Assumptions:} The graph $G(t)=(V,E(t),A(t))$ satisfies the following.
\begin{enumerate}
    \item For all $t\geq 0$ , the weight matrix $A(t)$ is doubly stochastic.
    \item There exists a positive scalar $\xi$ such that $[A(t)]_{ii}\geq \xi$ for all $i$ and $t\geq0$ , and $[A(t)]_{ij}\geq \xi$ if $[A(t)]_{ij}>0$.
    \item There exists an integer $T\geq1$ such that the graph $(V,E(s_T)\cup ...\cup E((s+1)T-1))$ is strongly connected for all $s\geq 0$.
\end{enumerate}

Once the robot $i$ model the network and constraints, g$^2$o \cite{kummerle2011g} optimize the provided k-ERPMG to find the best possible position estimations for connected $n$ robots in the form of optimized graph $G_o = (V_o, E_o, A_o)$. $V_o$ are the possible node positions concerning $i$, $E_o$ contains the possible distance range to other nodes in terms of weighted edges, and $A_o$ is an optimized adjacency matrix from $i$ to every other node. Algorithm \ref{alg: RLGO} describes the entire localization strategy for a given MRS, with a graph-based optimizer denoting the algorithm's three main iterative steps.

 \begin{algorithm}[t]
 \SetAlgoLined
  At given time interval $t$, for robot $i$\;
  Observe connected robots n\;
  Receive motion and observability information from connected robots\;
  Generate ERPMG from RPMG using Alg. \ref{alg: ERPMG}\;
  Check observability of n-robot systems using Alg. \ref{alg: OCheck}\;
  \For{ every sampling instance $t_k$, k = 0,1,2,..}{
      \For{each connected robot $R_j$, $j\neq i$ and j = 1,2,...,n }{
        Get previous odometry information\;
        Get previous relative positioning information\;
        Generate $n^k$ possible graph over estimation horizon\;
      }
    }
  Set constraints as Eq.~\eqref{eqn: constraints}\;
  Generate a time-varying network model for $n$ connected robots\;
  Solve the graph optimization problem using distributed solver \cite{kummerle2011g} over a predicted horizon\;
  \For{each connected robot $R_j$, $j\neq i$ and j = 1,2,...,n}{
   Extract position estimation from optimized graph: $X_{j,t+1} = (x_{j,t+1},y_{j,t+1},\phi_{j,t+1})$\;
  }
  \caption{Relative Localization Based on Graph Optimization}
  \label{alg: RLGO}
\end{algorithm}

\section{Theoretical Analysis}
Graph optimization-based relative pose localization has three submodules; Formation, Expansion, and Optimization. The formation depends on the number of robots $n$, which causes the generation of ERPMG $G=(V, E, A)$ and expansion relies on the number of possible position instances, $k$, which produces $n^k$ potential graphs to be optimized. As both initial steps can be serially done in polynomial time, one can deduce that the algorithm is polynomial and scalable with the number of robots. However, the optimization of $n^k$ possible graphs using the g$^2$o optimizer needs to be analyzed for its NP-hardness.

\textbf{NP-hardness:} We will show that the problem of finding optimizers is NP-hard in general. An example shows that L1-optimality is NP-hard for non-submodular energies.

Remember that if no two vertices of a graph $G=(V, E, A)$ are connected by an edge, the set $U$ of vertices is independent. The problem of finding the maximal independent set of vertices of an arbitrary graph is known to be NP-hard \cite{cormen2022introduction}.
Consider the following local costs as an example:
\begin{itemize}
	\item Give each vertex $v$ of label $l$ the cost $l_i$.
	\item For each edge with both vertices of label $l$, let the cost be $N = |V|+l$.
	\item Connect the cost of 0 to any other edge.
\end{itemize}
It is worth noting that if and only if the set $U=l_1(1)$ is independent, the maximum cost of any labeling $l$ is $N$. All labeling $l$ associated with an independent $U$ have a maximum cost of 1. Furthermore, the labeling $l$ is a strict minimizer when the number of cost 1 atom for $U$, which is $|V|-|U|$, is minimal, i.e., when the size of $U$ is maximal.
To put it another way, if we use the previously mentioned local cost assignments for a graph $G$, then $l$ is a strict minimizer if and only if $U:=l_1(1)$ is a maximal independent set of vertices. 
As a result, our problem, like the problem of finding the most extensive independent set of vertices, is NP-hard.
Thus, the optimization problem is proved to be NP-hard, and we also know that such graph problems can be verified in polynomial time \cite{cormen2022introduction} hence the problem is NP-complete.

\textbf{Convergence:} 
In order to control convergence, the g$^2$o method adds a damping factor and backup operations to Gauss-Newton, for which g$^2$o solves a damped version of the optimization function. A damping factor $\lambda$ is present here; the more significant it is, the smaller the $\lambda$ is. In the case of nonlinear surfaces, this helps control the step size. The g$^2$o algorithm's goal is to regulate the damping factor dynamically. The error of the new setup is tracked throughout each iteration. The following iteration is reduced if the new error is smaller than the prior one. If not, lambda is raised, and the solution is reversed. We refer to \cite{kummerle2011g} for a more thorough discussion of how the g$^2$o algorithm guarantees convergence.
\begin{figure*}[t]
    \centering
%\begin{subfigure}{\linewidth}
\centering
\hspace{-8mm}
\includegraphics[width=0.99\linewidth]{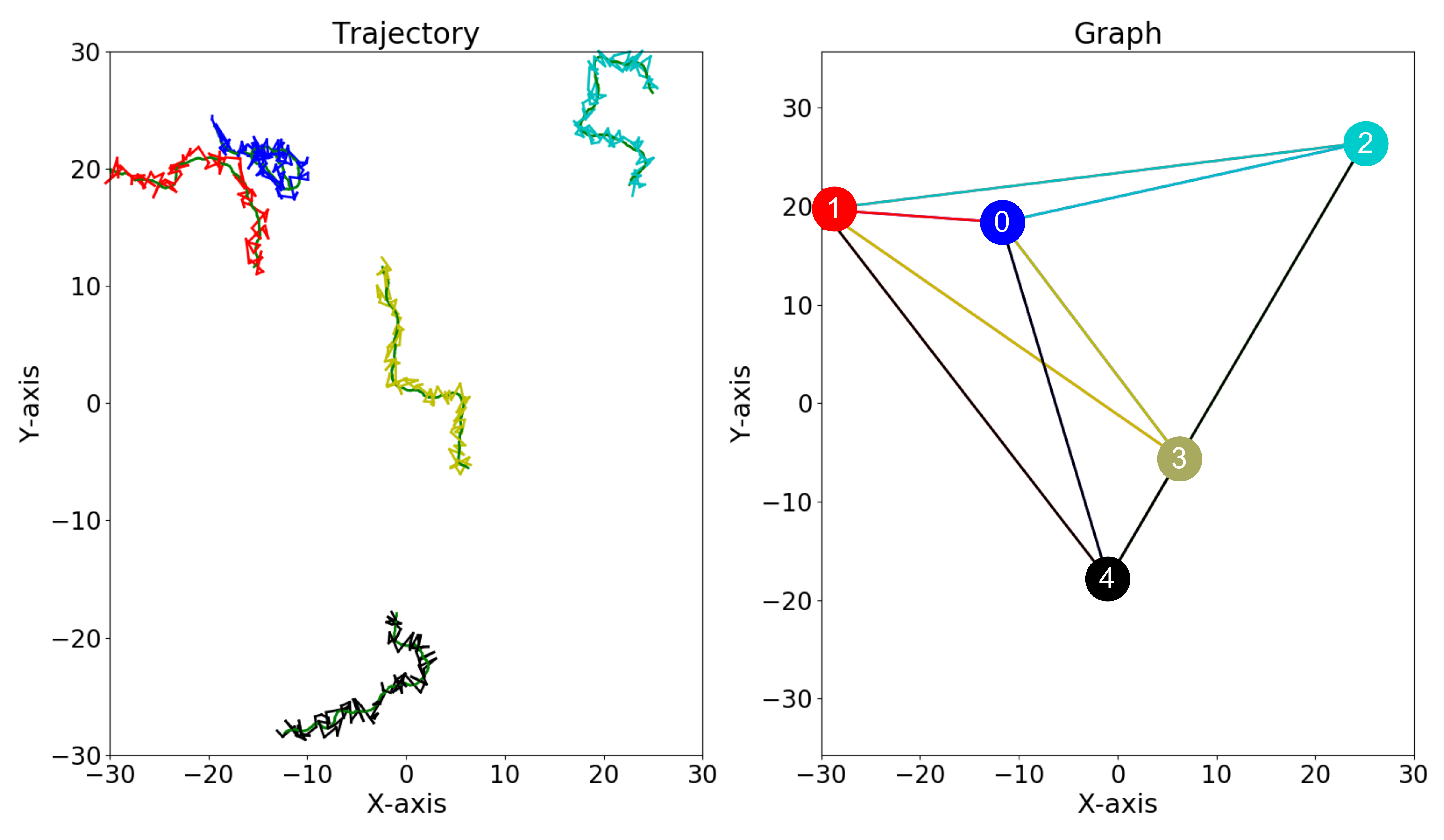}
%\end{subfigure}
\caption{Trajectories and the Optimized Graph. \textbf{Left}: Simulation with the ground truth (green) trajectory along with the colored predicted trajectory of a robot; \textbf{Right}: Robots as colored vertices and colored edges between connected robots with the respective color for optimized graphs. }
    \label{fig:traj_graph}
\end{figure*}

\begin{figure}[t]
    \centering
%\begin{subfigure}{\linewidth}
\centering
\includegraphics[width=0.61\linewidth]{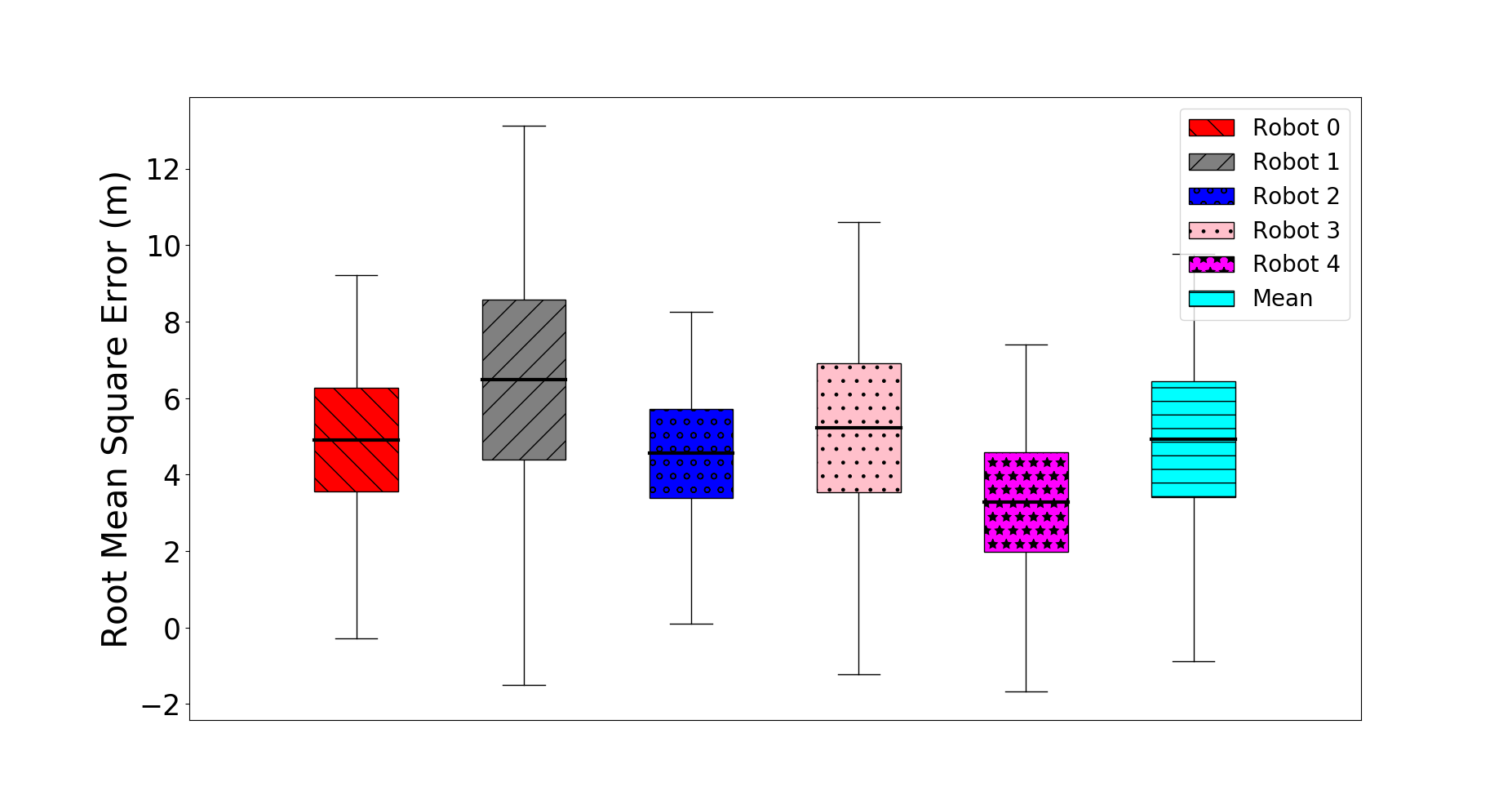}
\includegraphics[width=0.38\linewidth]{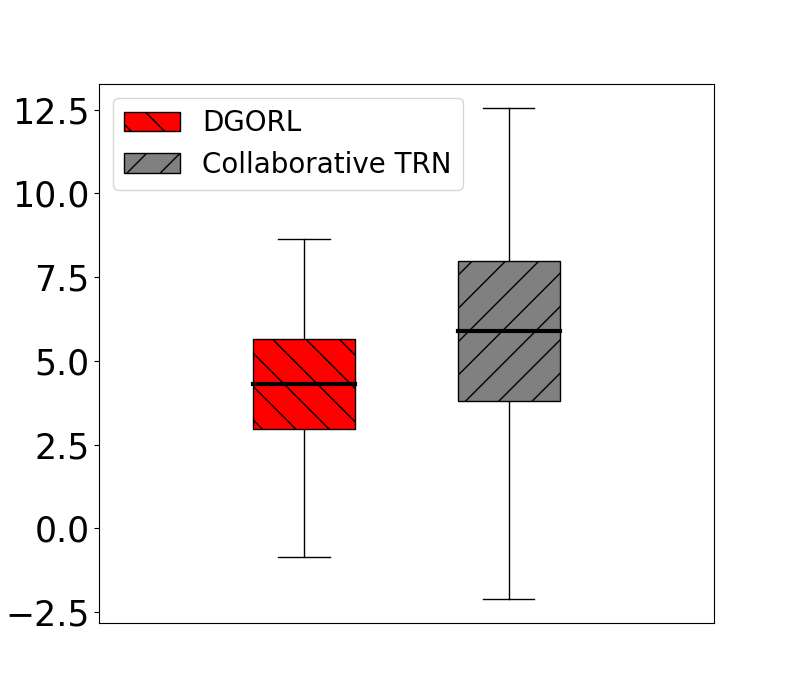}
%\end{subfigure}
\caption{DGORL performance of five robots in terms of Root Mean Squared Error (RMSE) in the 60m x 60m simulation workspace (\textbf{Left}). Comparison with Collaborative Terrain Relative Navigation (TRN) based on Covariance Intersection (\textbf{Right}).}
    \label{fig:results}
\end{figure}

\iffalse
\begin{figure}[t]
    \centering
%\begin{subfigure}{\linewidth}
\centering
\includegraphics[width=0.98\linewidth]{mrl_graph_comp.png}
%\end{subfigure}
\caption{Comparison graph of average performance of Distributed relative localization of five robots in terms of Root Mean Squared Error with Collaborative Terrain Relative Navigation (TRN) based on Covariance Intersection}
    \label{fig:comp}
\end{figure}
\fi

\section{Simulation Experiments and Results}
\label{sec:results}
We performed extensive simulation experiments in a 60 x 60 meters bounded region under certain conditions to analyze the algorithm. We have set up the MRS as discussed in Sec.~\ref{sec:problem}, where each robot shares inertial measurement and RSSI with connected robots. Each robot in the MRS performs relative localization based on its frame of reference. we can calculate weights of edges $e_{ij}$ in $G=(V,E)$ as a range $d$ the  between $R_i$ and $R_j$ using path loss model using perceived RSSI:
\begin{equation} 
d = 10 ^ \frac{A - RSSI}{10n}
\end{equation}
Here, $n$ denotes the signal propagation exponent, which varies between 2 (free space) and 6 (complex indoor environment), $d$ denotes the distance between robots $i$ and $j$, and $A$ denotes received signal strength at a reference distance of one meter.
On every iteration, $r_i$ generates ERPMG based on its observation and performs observability analysis (at least four nodes connected, based on the observability Checking in Alg. \ref{alg: OCheck}). Later, it will expand ERPMG and generate input for the g$^2$o optimizer. 
%MOVED to Introduction section
%General Graph Optimization (g$^2$o), an open-source graph-based framework to handle the nonlinear error problems, was used to optimize global measurement pose using the initial global measurement poses and local relative pose constraints. 

The initial global measuring posture, the local relative pose limitations, and its information matrix make up g$^2$o's input. Once obtained an optimized graph, $r_i$ estimated relative positions from vertex poses. Although in MRS, each robot performs localization distributedly. To measure error, we have captured initial positions on a global scale and compared predicted trajectories with ground truth as RMSE in meters. We used range sensors to compare our strategy to Terrain Relative Navigation (TRN) from \cite{wiktor2020ICRA}.

\paragraph*{Localization Accuracy} 
We performed experiments with five robots driving them on a random walk for 100 iterations, and obtained consistent results through 10 repetitive trials under identical conditions. Fig.~\ref{fig:traj_graph} visualizes the predicted trajectories and connected graphs for experimentation. Results in Fig.~\ref{fig:results} have validated the claim about the accuracy of DGORL in a larger space, with 8\% localization error.

Furthermore, experimentation was carried out in the same simulated workspace area to evaluate the localization accuracy of DGORL in contrast to TRN and to confirm its applicability in more significant scenarios. We performed ten attempts using identical simulation conditions and obtained the RMSE in meters to assess the localization accuracy. Results in Fig.~\ref{fig:results}, which demonstrate a 23\% percent improvement in localization accuracy over TRN, have supported the assertion that DGORL is viable in larger contexts. Furthermore, the absolute localization error of 4.2 meters for a $3600m^2$ zone with five robots in a standalone view of localization of DGORL is highly encouraging. Still, it needs further analysis to understand the limitations and sufficient conditions. Nevertheless, the results showcase the viability of DGORL for MRS applications.

\paragraph*{Computational Demand}
The mean optimization time (referring to the time taken by the solver) and the mean CPU usage of the whole process are $10.2\pm 2.7$ms and \textcolor{black}{$143 \pm 49\%$, } respectively. The results are evaluated for five robots from 10 runs over 100 iterations. The computational complexity of covariance intersection-based TRN is close to DGORL, which further highlights the efficiency of DGORL. It is worth noticing that the performance of DGORL provides evidence for its practical use in small resource-constrained robots (i.e., equipped with tiny computers, e.g., Raspberry Pi zero with Quad Core running at 1 GHz). Hence, DGORL can be used as an efficient relative localization algorithm for most multi-robot systems, including swarm robotics.

\section{Conclusion}
Many estimations, planning, and optimum control challenges in robotics have a basis for optimization problems. In most optimization problems, the goal to be maximized or minimized is made up of multiple local components or terms. That is, they only rely on a tiny portion of the variables. We use graph theory to solve a well-defined multi-robot relative localization problem. Graph generation, expansion through transition, and optimization are the three fundamental components of our novel technique. We first estimated a relative pose measurement graph using received signal intensity between connected robots. Later, we used a motion model to build graph expansions and a polynomial-time graph optimizer to find an optimized pose estimation graph from which a robot extracts relative positions for connected robots. Finally, we analyzed the algorithm through simulations to validate the practicality and accuracy in a large workspace. Furthermore, DGORL delineates high localization accuracy than appropriate methods while maintaining a comparable optimization time, which further backs the efficiency and accuracy of DGORL. In the future, we will perform hardware experimentation to substantiate the applicability of graph optimization in the real world.

\def\bibfont{\normalfont\small}
\bibliography{ref}
\bibliographystyle{IEEEtran}

\end{document}